%% file: 0_main.tex
\documentclass[11pt, a4paper]{article}
\usepackage[utf8]{inputenc}
\usepackage{authblk}
\usepackage{pdfpages}
\usepackage{outlines}
\usepackage[autostyle]{csquotes}
\usepackage{array}
\usepackage{caption}
\usepackage{rotating}
\usepackage{color,soul}
\setstcolor{red}

\usepackage{placeins}
 \usepackage{hyperref}  
 \usepackage[numbers]{natbib}
 
\hypersetup{
    colorlinks=true,
    linkcolor=blue,
    filecolor=magenta,      
    urlcolor=cyan,
    citecolor=blue,
}

\title{Towards an Explanation Space to Align Humans and Explainable-AI Teamwork}

\author[1*]{Garrick Cabour}
\author[1**]{Andrés Morales-Forero}
\author[2]{Élise Ledoux}
\author[1]{Samuel Bassetto}
\affil[1]{Mathematics and industrial engineering (MAGI), Montreal University (Polytechnic), Montreal, Canada}
\affil[2]{Department of Physical Activity, Université du Québec à Montréal, Montreal, Canada}
\affil[*]{Corresponding author: \tt{garrick.cabour@polymtl.ca}}
\affil[**]{\tt{andres.moralesforero@polymtl.ca}}

\date{} 

\providecommand{\keywords}[1]{\noindent\textbf{\textit{\small Keywords---}} \small #1}

\emergencystretch=\maxdimen
\begin{document}

\maketitle

\begin{abstract} 
\noindent Providing meaningful and actionable explanations to end-users is a fundamental prerequisite for implementing explainable intelligent systems in the real world. Explainability is a situated interaction between a user and the AI system rather than being static design principles. The content of explanations is context-dependent and must be defined by evidence about the user and its context. This paper seeks to operationalize this concept by proposing a formative architecture that defines the explanation space from a user-inspired perspective. The architecture comprises five intertwined components to outline explanation requirements for a task: (1) the end-users mental models, (2) the end-users cognitive process, (3) the user interface, (4) the human-explainer agent, and the (5) agent process. We first define each component of the architecture. Then we present the Abstracted Explanation Space, a modeling tool that aggregates the architecture's components to support designers in systematically aligning explanations with the end-users work practices, needs, and goals. It guides the specifications of what needs to be explained (content - end-users mental model), why this explanation is necessary (context - end-users cognitive process), to delimit how to explain it (format - human-explainer agent and user interface), and when should the explanations be given. We then exemplify the tool's use in an ongoing case study in the aircraft maintenance domain. Finally, we discuss possible contributions of the tool, known limitations/areas for improvement, and future work to be done.  
\end{abstract}


\section{Introduction}

\input{1_intro}

\section{Background and Motivations}

\input{2_background}

\section{Proposal: User-Centered XAI Framework}

\input{3_proposal}

\section{Case Study: Aerospace Industrial Inspection}

\input{4_casestudy}

\section{Discussion \& Conclusion}

\input{5_dicussion}

\noindent\textbf{Acknowledgments.} This research is supported by the Consortium for Research and Innovation in Aerospace in Québec (CRIAQ), funded by Mitacs accelerate program (contract \#Manu-1712 - IT11797). The findings and conclusions in this report are those of the authors.\\

\noindent\textbf{Conflict of interest.} The authors declare that there is no conflict of interest.
\FloatBarrier

\bibliographystyle{ieeetr}

\bibliography{bib}

\end{document}

%% file: 1_intro.tex
The design of AI-powered systems has been driven primarily by technical aspects and considerations \citep{vasey2021, cabitza2019}. However, as computer systems are increasingly integrated into work environments, new forms of Human-AI collaboration emerge \citep{jiao_towards_2020, peeters2021}. Achieving business goals in partnership with artificial agents raises new social, technical, and operational challenges \citep{cabitza2019, cabour2021, cutillo2020, johnson2017}. The transparency of AI models is a challenge that receives increasing attention from the scientific community \citep{doshi-velez2017}. 
Enabling Human-AI collaborative capabilities requires an understanding of \textit{automation's inner workings} \citep{rajabiyazdi2020}. A sub-field of Safe AI - eXplainable AI (XAI) - is pursuing this objective \citep{mueller2019}. This paradigm aims to make complex machine reasoning interpretable for human beings to avoid blind decision-making issues \citep{mueller2019, mueller2021, hoffman2018}. Effective communication of the decisions and actions taken by automation is a prerequisite to allowing conjunctive task accomplishment in Human-AI teams \citep{rajabiyazdi2020, johnson2021}. Some authors have thus argued that an explanation is a situated interaction between users and the AI system, rather than a \textit{property of statements} \citep{hoffman2018, miller2019, ribera2019, doshi-velez2017, mueller2021}. 
\blockquote{\textit{What counts as an explanation depends on what the user needs, what knowledge the user already has, and especially the user’s goals. This leads to a consideration of function and context of the AI system (software, algorithm, tool). That is, why does a given user need an explanation?} \citep{hoffman2018}}
Following the footsteps of interaction design, providing users with meaningful explanations encompasses many human, cultural, and organizational context-dependent factors \citep{cutillo2020, cabitza2019, miller2019}. 
Thereby, designing explanations is a multifaceted endeavor that integrates technical (XAI capabilities, model selection), humans (users' needs, goals, and information-processing abilities), and socio-technical (regulations, rules, procedures, and actors involved) \citep{cabitza2019, selbst2019, cabour2021} perspectives. 
Many researchers have proposed a complete list of requirements that fall into the technical aspects of XAI .e.g., delivering human-readable output \citep{mueller2021, ribera2019}. While these criteria are essential prerequisites, they do not guarantee that the explanations align with users' goals, needs, and operating contexts \citep{mueller2021}. Only task-relevant information should be disclosed to the users to avoid putting them in a state of cognitive overload \citep{chen2014a, rajabiyazdi2020}.\\


What is missing from the current literature is an integrated framework that links user studies to XAI technical development, keeping people \textit{at the center of the design and evaluation process} of explanations  \citep{vasey2021}. 
This paper proposes a user-inspired formative guideline that supports the design of explanations to address the gap mentioned above. We have identified five interrelated dimensions that constitute the explanation space. These dimensions support a user-centered view to design tailored explanations by linking user needs with the XAI's technical capabilities/possibilities. We illustrate the use of the framework in an ongoing case study of XAI development in the field of aviation maintenance.\\

This paper is organized as follows. We begin by providing a concise and critical overview of current XAI frameworks and methods for studying users in the wild. We then present our formative framework, detailing the features that comprise it. We then exemplify how we currently apply this framework in an industrial case study. Finally, we discuss practical benefits, limitations, and known areas of improvement.

%% file: 2_background.tex
The literature on Explainable AI is evolving rapidly. Here we briefly review the most relevant work for the sections. 

\subsection{Why Explain?}

The incremental implementation of AI in the workplace will result in more and more Human-AI interaction \citep{seeber2020m}. Technological capabilities now support human performance in high-level cognitive tasks: decision-making, sense-making, planning, etc. One of the main components to support human performance - and simultaneously the multifaceted concept of trust - is the intelligibility of the system's actions,  reasoning,  and potential mistakes \citep{matthews2021, chen2014a}. This challenge is not new, as previous studies on experts' systems revealed the issues associated with the system's reasoning opacity \citep{mueller2019}. Among them, problems related to the interpretability of actions, information processing, and decision choices resulted in decreased performance, missus, and rejection of the technology by the users \citep{clancey1983, moore1988, zouinar2020}. This opacity indeed engendered uncertainty, disturbing the process tasks accomplishment for which the expert system was involved \citep{roth1987, zouinar2020}. The problems encountered during the expert system's design are even more topical for contemporary AI-powered solutions as new ML solutions, such as deep neural networks, are less understandable and more opaque than IF-THEN rules \citep{miller2019, lewis2021deep}. In addition, AI-based solutions are being implemented in safety-critical domains, such as healthcare and aviation\footnote{https://aiir.nl/} \citep{habli2020, shmelova2020}. \\

Deep Learning has revolutionized the technological industry generating algorithms that have surpassed the human precision level in certain instances \citep{mnih2015human}. These algorithms are generally made up of recursive combinations of hundreds of non-linear functions that make them powerful but incomprehensible. Adopting these technologies in a context such as justice, transport, and healthcare has harmed society \citep{marcus2019rebooting}. The decision mechanism of high-performance ML models can be based on hidden spurious correlations \citep{lapuschkin2019}, leading their users to make mistakes. XAI would help to judge the decisions made by the automated system and understand and prevent misuses of their results in a given context. For instance, understandable decisions would enable physicians to correct any decision made by an AI system \citep{akatsuka_illuminating_2019} and prevent a false diagnostic; interpretability is needed because military decision-makers must be able to justify their decisions \citep{tomsett_why_2018}; judges or policymakers refine, criticize and trust in their decisions \citep{lakkaraju2016interpretable}; explainable systems would help the inspectors to analyze aircraft components; driverless cars would provide users of helpful information to make decisions when the human common sense is required. Therefore, XAI is becoming essential for certain privates companies and officially mandatory for several countries: European Ethics Guidelines for Trustworthy AI \citep{pekka2018european}; The Canadian Guide of Principles for effective and ethical use of AI; OECD Principles on AI \citep{yeung2020recommendation}.\\



Besides the above-mentioned, XAI would allow finding adversarial examples. An observation that can cause a classifier to err is known as an adverse observation. Finding adversarial examples is essential for the safety of intelligent critical systems. Although there are methods for automatic generation of adverse examples \citep{xiao_generating_2019}, this is still a growing research field. Anomalies, novelties, rare events, or outliers are often challenging to find, even more so if they are not in these models' training data set, some anomaly types are unknown \citep{morales2019case}. Understanding the mechanism within these systems can allow finding weak points settings and make their algorithms more robust in a real-world context \citep{fidel_when_2019,tramer_adversarial_2019}.


\subsection{What and How to Explain?} \label{whatandhow}

XAI aims to develop methods that can explain their own decisions or other AI model's decisions: XAI approaches might be either (1) \textit{inherently interpretable models} or (2) \textit{post-hoc methods}. Inherently interpretable models (such as regression models, logistic models, Bayesian models, or decision trees) have an understandable architecture that makes them, to some extent, human interpretable. On the contrary, post-hoc methods are applied after the black box model has been already trained. Some of those methods might require full or partial access to the internal black-box mechanism, and others would only require query access (outputs from specific inputs).\\

Although inherently interpretable models have been used in contexts such as health \citep{mor-yosef_ranking_1990}, or education \citep{kobrin2011investigation}, more complex models have outperformed them. These interpretable models can provide more faithful interpretations \citep{rudin2019stop}; nevertheless post-hoc methods, especially \textit{agnostics approaches}, might be more in line with the unbridled development of AI technologies. Agnostics explainers belong to a post-hoc methods family that only requires query access to the black box, e.g. SHAP \citep{lundberg2017unified} ;LIME \citep{ribeiro2016should}; G-REX \citep{konig2008g}; ASTRID \citep{henelius2017interpreting}. They are attractive since they ignore the underline structure of the target model, allowing a broader spectrum of applications. In some cases disclosing the system's inner mechanism can make it vulnerable to attacks or gamification \citep{rathi2019generating}. Systems gamification may be useful in certain circumstances \citep{avserivskis2014gamification}, but counterproductive in others \citep{shahri2014towards}.\\

Nevertheless, before considering how explainability is obtained, a careful examination of user needs and context should inform what needs to be explained. The answer to such a question depends on context-dependent and context-independent criteria. Indeed, an explanation may focus on understanding the system's reasoning process (context-independent) or a particular decision (context-dependent). \cite{doshi-velez2017} distinguishes these two types of focus as \textit{global} (model of system reasoning and learning) and \textit{local} (justification). Justifying a decision is not always required and can be guided by eliciting end-users needs and decision-making context \citep{mueller2021, naiseh2020}. For instance, \cite{naiseh2020} formalized four significant factors that drive explanation requirements and personalization. They include the 1) users' goals; 2) their cognitive behavior (the way people process information and make a decision); 3) criticality or importance of the decision; and 4) compliance with AI regulations. \cite{mueller2021} also argue in that direction and add other explanation requirements such as time-constraint requirements. Indeed, the explanations required may differ at different points in the process. They argue that effective XAI systems that help the users achieve their goals must follow human-centered design principles. Users-based researches can build on the knowledge that provides guidelines for defining local and global explanation requirements: Is the output sensible and validated by users? What information do the users need to know to fulfill work requirements? How the model made such a prediction?\\


In addition to a \textit{local} and \textit{global} focus, the format of explanations can also be contrastive or counterfactual. These kinds of explanations are pretty similar and are based on-example explanations where the user may interact, to some extent, with the machine, i.e., there may be a complete user-machine communication. The user can access different examples, decide which might be the most reasonable output, or wonder why this answer is not the other one? What would be the result if something changes in the model or the input? This interaction with the system might improve its perceived safety; the user experience might support system accuracy  \citep{morales2021un}. However, this explanation space is defined for the actors' requirements, so the actors are the ones to decide which questions are needed to answer \citep{mohseni_multidisciplinary_2020}. Therefore, all the actors' roles, needs, and knowledge must be considered in designing explainable intelligent systems.\\ 

User-friendly interfaces must also be provided for the system. The system's functionality may depend on how the machine communicates with the user \citep{mohseni_multidisciplinary_2020}. Explanations can be visual, textual, or more complex explanations, and the type of explanations depends on the users and its context.  While some users require partial explanations focusing on relevant features, others may require the complete portrait of features and reasoning behind the decision \citep{naiseh2020, lewis2021deep}. Moreover, some critical systems might also be required to provide specific explanations to official regulatory agencies, so their mandates must also be fulfilled.


\subsection{Inferring Explanation Needs Through User Studies} \label{usercentered}

Linking user research to XAI calibration is critical for providing relevant local and global explanations to end-users\footnote{The word end-user refers to the actors in direct interaction with the technical device}. Only a few papers have tackled this issue, and there is currently a severe gap in obtaining and using end-users insights in the design of XAI systems \citep{mueller2021, naiseh2020, sanneman2020}. This section presents relevant approaches to inferring end-user needs through ethnographic-based studies.\\

Ethnographic-based studies aim to understand how people behave in real-world contexts and why they behave the way they do \citep{crabtree2012, bisantz2015}. Ethnographers immerse themselves in the users' natural environment (\textit{in the wild}) to gain an in-depth understanding of people's actions, intentions, and decisions \citep{crabtree2012}. Using ethnographic fieldwork techniques, field researchers elicit and derive valuable information about the users' goals, needs, tasks, knowledge, and the work context in which they operate \citep{bisantz2015}. The purpose of doing so is generally design-oriented, meaning that ethnographers conduct field research to support a higher-level motive of artifacts/work design and not \textit{doing ethnography per se} \citep{crabtree2012}.

There are many analytical frameworks for analyzing human (user) behavior in real work situations: the \textit{naturalistic decision-making} (NDM) \citep{zsambok2014, bisantz2015, klein2016}, \textit{Cognitive Work Analysis} (CWA) \citep{vicente1999, bisantz2015}, \textit{Ergonomic Work Analysis} (EWA) \citep{st-vincent2014, salembier2021}. Each of these frameworks enables the analyst to provide detailed descriptions of the work conducted by subject-matter experts (SMEs), especially the socio-cognitive dimension (e.g., problem-solving strategies, decision-making processes). Also, NDM, CWA, and EWA recognize the importance of context as factors modulating work realization, and thus the cognition and possible course of action of end-users ("field of possibilities") \citep{endsley2007, bisantz2015, st-vincent2014}. For example, industrial environments lean on high-quality standards to meet customer requirements and minimize production costs, ultimately determining the rules that operators must apply to evaluate the manufactured parts. These work domain parameters may motivate the need for additional explanations. Thereby, each of the presented frameworks relies on a system approach that involves two complementary layers of analysis to understand users' activity in complex environments \citep{bisantz2015, st-vincent2014}:

\begin{outline}[enumerate]
\1 Work Domain Characteristics (the "why"): "focus on the \emph{domain-driven factors} that shape, support and constraint human performance and behavior" \citep{bisantz2015}.
\2 Task requirements: \textit{expected outcomes in terms of quantity and quality of work}; \textit{degree of instructions' clarity} \citep{st-vincent2014} e.g., ill-defined instructions
\2 Organizational settings: organization of work, teamwork \& interdependencies among workers, task allocation, work system processes, employees' training. 
\2 Physical environment: workstations, equipment and materials
\2 Socio-cultural environment: hierarchical relationships, decision-making operational leeway, institution rules and legislation  
\1 User Activity (the "what and "how"): focus on the situated aspect of SMEs' work encompassing problem-solving strategies, meaning-making, and decision-making processes, knowledge applied,  and skills required \citep{cabour2021} \textit{that allow them to operate successfully in the domain} \citep{bisantz2015}. The goal is to deepen the analysis of expertise to uncover how operators deal with the domain's characteristics (such as work requirements and institution rules), the inherent variability of processes, the decentralization of work, and the dynamic nature of contemporary work environments \citep{bisantz2015, cabour2021a, goh2020}.
\end{outline}


Field studies are accompanied by various data collection techniques to elicit relevant content on work domain characteristics and user activity/mental models\footnote{User understanding and representation of a process, a phenomenon, or a system \citep{hoffman2018}}. The techniques commonly used involve observations, interviews, focus groups, process mapping, card sorting, fieldwork experimentation with SMEs \citep{shadbolt2015, hoffman2018, bisantz2015, st-vincent2014, milton2007}. Researchers vary their data collection methods to avoid bias and ensure rigorous methodology \citep{wilson2015}. This variation also aims to elicit knowledge that would not be possible to obtain with a single method. For example, think-aloud protocols (concurrent verbalization when realizing a task) and retrospection interviews with textual, audio, or video field data are well-equipped to deepen the \textit{in-situ} observations. Analysts may be able to understand the cognitive work requirements, information used, and the problem-solving strategies of SMEs in real-time; and ii) to support a "scaffolding" construction of users' mental models\footnote{See \cite{hoffman2018,wilson2015, milton2007} for a complete description of existing data collection methods and techniques} \citep{hoffman2018}. 
In the case of XAI, the output of these field studies is to provide guidance in the specification of explanation needs (whether local or global).

%% file: 3_proposal.tex
\subsection{Methodical considerations}
\begin{figure}[h!]
	\centering
	\includegraphics[width=1\textwidth]{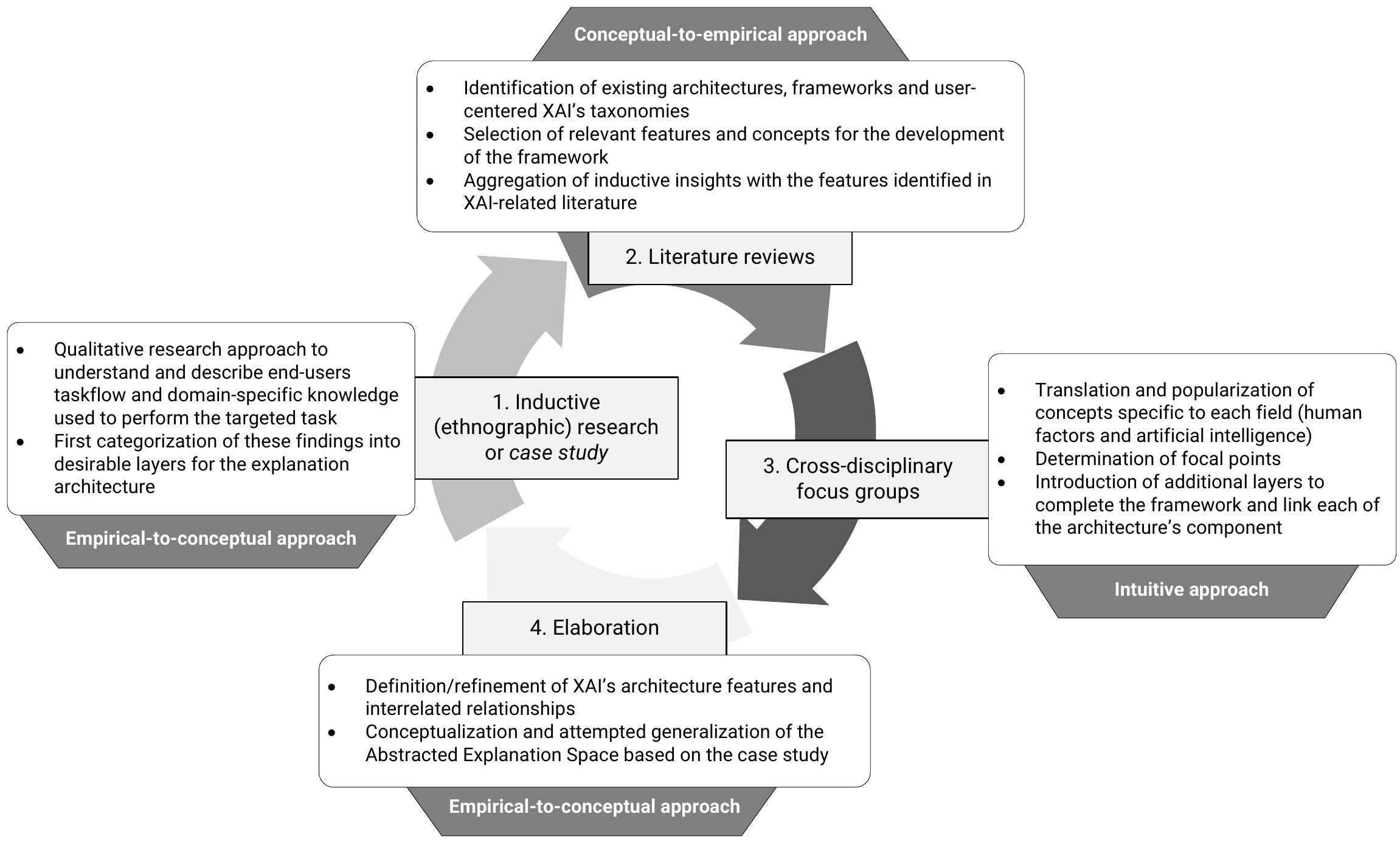}
	\caption{Overview of our methodical approach to elaborate the framework}
	\label{fig:metho}
\end{figure}

The proposed methodology for conceptualizing and elaborating a user-centered XAI framework consisted of four iterative steps depicted in Figure \ref{fig:metho}. Considering that user-centered XAI's research is a cross-disciplinary and relatively new topic, we combined different approaches to propose the user-centered framework. Our methodological development is inspired by the taxonomy design phases proposed by \cite{nickerson2013} (especially for the choice of different inductive/deductive/intuitive approaches) and by the step-by-step process proposed by \cite{mcmeekin2020} for framework development. An empirical-to-conceptual approach is defined by a set of known features that the researchers \textit{wish to classify} or aggregate \citep{nickerson2013}. These sets of features are elicited from inductive research in our case. In contrast, in a conceptual-to-empirical approach, the \textit{researcher begins by conceptualizing the dimensions of the taxonomy without examining actual objects} \citep{nickerson2013}. Finally, an intuitive approach is characterized by researchers' \textit{ad hoc} endeavors to develop an \textit{artifact} (here, some aspects of the framework) based on their current understanding and \textit{perceptions of what makes sense} \citep{nickerson2013}.

We first determined how we can qualify end-users explanation needs based on a current qualitative case study. Using an empirical-to-conceptual approach, we evaluated the framework's core features related to the human component of the architecture.

Based on dual-objective literature reviews on XAI's context-dependent and context-independent considerations, we deduced the most relevant features for the development of our framework. We then aggregated the evidence-based knowledge obtained during the inductive research to fit the identified features (conceptual-to-empirical approach).

In the third phase, given the literature's current limitations and lack of cross-disciplinary endeavors for XAI, we used an" intuitive approach" to aggregate, translate or propose additional relevant concepts to complete our framework \citep{nickerson2013}. This step required several discussions and focus groups on identifying focal points and introducing additional features for the framework. In addition, the interrelation among the framework's features also followed an intuitive approach.

Finally, we refined our architecture in the last step. This refinement follows an iterative process that may result in the need for additional work in previous phases. Based on the explanation space we draw from our case study, we conceptualized and generalized a context-independent framework with its associated formative tool: the Abstracted Explanation Space. The tool's development thereby followed an empirical-to-conceptual development.

\subsection{User-centered XAI framework}

XAI specialists argue for a more human-centered perspective to address local and global explainability \citep{hoffman2018, ribera2019, doshi-velez2017, selbst2019}. Such an approach would capture the essence of end-users work context to develop task-specific explanations that align with their goals. 
The proposed framework is consistent with this line of thinking. Our approach is motivated by a systemic approach that considers the operating context as a determining element in the design choices of the XAI. Before presenting it, we wish to clarify its scope and limitations. The formative tool (Abstracted Explanation Space, see \ref{aes}) supporting the deployment of the framework that aims to guide the design of" tailored" explanations for real-work situations, particularly identifying and validating the relevant content that users need to perform a specific task. Therefore, aspects concerning the calibration of ML models \textit{per se} are not included in this document, nor are the effects of explanations on users (e.g., trust). Motivations stem from a persistent gap between ethnographic analysis (or user studies) and system design, reinforced by new AI-powered system design methodologies \citep{johnson2017, cabitza2019}. Because XAI relies on context-dependent criteria, designers need tools that help them identify and use contextual insights to design tailored explanations and select the appropriate presentation format (based on business, work, and end-users requirements). The goal is to guide designers on what, why, how, and when information about the tasks or rationale of the automated system should be disclosed \citep{rajabiyazdi2020}. Drawing on ethnographic approaches, it starts by gathering business and work requirements to define what should be explained from the users' perspective. The knowledge gathered from ethnographic inquiries is then used to define the appropriate ML model(s) that provide the required explanations. Five features that shape the human-XAI interaction were identified, ultimately outlining the explanation space for a \emph{specified task} (Table \ref{table:xai}).

\begin{table}[h]
\centering
\small
\caption{XAI features}
\label{table:xai}
\begin{tabular}{|m{3cm} |m{8.75cm}|}
\hline
\textbf{Features} & \textbf{Content} \\ [0.5 ex]
\hline
End-users mental model & A simplified representation that a person has of the
structure and functioning of a system, concept, process, or problem to be solved. It encompasses different forms of background knowledge, whether explicit (domain concept and principles), declarative, procedural, and contextual \citep{hoffman2018}.\\
\hline
End-users cognitive process & Mental steps/actions that a user undertakes to achieve a
goal (usually represented in a \textit{flowchart}). Procedures can specify part of these steps, but users' studies must draw a complete picture.\\
\hline
User Interface & Materialization of the interaction where informational requirements (content, format and timing of presentation) have been carefully studied to meet end-users needs (mental model and cognitive process).\\
\hline
Human-Explainer Agent & This agent provides meaningful explanations to the user of relevant classification made by the Classifier Agent.  It is a set of algorithms that interpret the classifier's decisions according to the objectives pursued by the users.\\ 
\hline
Classifier Agent & Set of ML models that automatically perform the classifications required for the task.\\
\hline
\end{tabular}
\end{table}

\subsubsection{Targeted Task and Socio-Technical Context}

Our approach is motivated by a system perspective that considers the operating context as a determining component of XAI design choices. Before working on the XAI technical system, it is necessary to understand the task's context.  The analyst responsible for this work package must collect data to understand the purpose of the work system and the role of end-users in it. "\textit{What is the end-user trying to achieve?}" To do so, he/she must uncover the domain-driven factors/constraints that shape human performance to understand and represent the end-users' work activity adequately."\textit{How and why is he/she working that way}"? (See \ref{usercentered} for theoretical details and practical considerations).\\

This step fulfills two goals. The first one is to determine automation opportunities in terms of task and explanation requirements. The second is to investigate how the socio-technical context (rules, regulations, and laws) constraints the realization of the work. Indeed, the design of a reliable system depends on its ability to respect the constraints imposed by the work domain. In addition, end-users might expect explanatory content regarding domain-driven constraints (e.g., expected quality outcomes and task success criteria).

\subsubsection{End-Users Mental Models}
After defining the task to be explained and its associated domain-driven constraints, the analyst must uncover how people make decisions based on their representation of the surrounding environment. This mental representation is often called a mental model in cognitive science \citep{friedman2018, hoffman2018}. It comprises an organized set of knowledge/concepts that structure the relevant aspects of a process, system, or phenomenon (with causal relationships between concepts) \citep{friedman2018}. The analyst aims to elicit end-users mental models for the targeted task using ethnographic fieldwork methods \citep{elsawah2015, hoffman2018}. The focus should be on understanding the cognitive work done by end-users and the relevant information/knowledge used. Second, identifying the why and how this knowledge is processed according to the objectives pursued and the domain-driven constraints.

\subsubsection{End-Users Cognitive Process}
The end user's cognitive process represents the decomposition of their cognitive task flow (or mental actions) for the targeted task. Depending on the complexity, the analyst can choose a coarse or finer layer of description.  This step can usually be represented with a hierarchical \textit{flowchart} or \textit{decision tree} \citep{shepherd2015, shadbolt2015}. The End-User Cognitive Process is linked to the Mental Models, as these contain the relevant information/knowledge they use for each operation. 

\subsubsection{User Interface}

The interface is the channel between the machine and the end-users. The fit between the end user's needs and the user interface is determined by a clear understanding of the end user's cognitive process, goals, and associated mental models. Indeed, understanding the user's context provides empirical data to define the explanations that should be disclosed: does the interface display the necessary explanations that enable the user to fulfill its operational goals (context-dependent criteria)? Has the representation format been designed to allow easy understanding of the system's relevant decision or reasoning features (context-independent criteria)? Do the explanations keep the user in the control loop while supporting their cognitive process (context-dependent criteria)?

On the other hand, the interface embodies the interaction between the explainable model(s) and the Human-Explainer agent. These explanations are exhibited in text, images, figures, or a combination of these three elements. They can be attribute-based, concept-based, contrastive, or counterfactual (See \ref{whatandhow} for theoretical details). They depend on the end-users needs for a specific task/operation and the capability of the Human-Explainer Agent. 

\subsubsection{Human-Explainer Agent} \label{humanexplainer}

The Human-Explainer Agent is a set of algorithms that explain the relevant decisions made by the (Classifier) Agent from the perspective of the end-user goals and tasks. The Human-Explainer agent interacts with the end-user through the User Interface. Explanations follow the three complementary layers from the Situation Awareness Agent-based (SAT) Transparency model \citep{chen2014a, chen2014b, sanneman2020} and are strongly inspired by the psychological concepts that characterize a good explanation for human beings \citep{chen2014a, chen2014b, sanneman2020, preece2018, miller2019}:
\begin{itemize}
\item XAI1 Perception ("What?"): concerns the relevant content that the user should perceive to pursue his goals or \textit{accomplish a task in correspondence with the automation} \citep{rajabiyazdi2020}. It is directly linked with the user goals, the knowledge he already has, and the knowledge he needs \citep{hoffman2018, mueller2021}. Examples of guiding questions are: What decision should be explained to the user during this operation? What part of the model should be interpretable? What aspect of the situation is relevant to the user? Is the decision sensitive, critical, or somehow important in the work context? 
\item XAI2 Comprehension ("Why?"): relates to the understanding of the Agent's rational model that led to this outcome (or causality).  Depending on the task and user needs, the explanations may focus on specific features of the situation or provide a complete description. Examples of guiding questions are: Why does the system provide this output? What are the reasons behind its decision? What are the governing rules, criteria, and domain-driven factors that influenced the decision? Should explanations be systematically displayed, or are they events, triggers, or cues that activate the need for explanations \citep{mueller2021}? 
\item XAI3 Projection ("What if?"): relates to the choices made by the Agent with other similar courses of actions by providing contrastive and/or counterfactual explanations \citep{miller2019}. Examples of guiding questions are: \textit{Why this decision was taken as opposed to other ones} \citep{lewis2021deep}? How would the system react to a slight change in the input? What happens if we adjust the X features/parameters of the image with Z?
\end{itemize}
These layered explanations are linked together and form an \textit{explanation object} that the user can access by a query \citep{preece2018}. The human may ask for finer details on AI actions/decisions by navigating through the different layers based on situational needs. Thus, each essential operation comprises an \textit{explanation object} that provides tailored explanations on agent action (" what"), agent decision (" why"), and counterfactual or contrastive explanations (" what if"; "how"). 

\subsubsection{Classifier Agent}

The Classifier Agent is a set of ML models that automatically performs the classifications required for the system. These ML classifiers can be either inherently interpretable or black boxes or both.  Each of their classifications will be interpreted by the Human-Explainer Agent.

\subsection{Abstracted Explanation Space} \label{aes}

\begin{figure}[h!]
	\centering
	\includegraphics[width=1\textwidth]{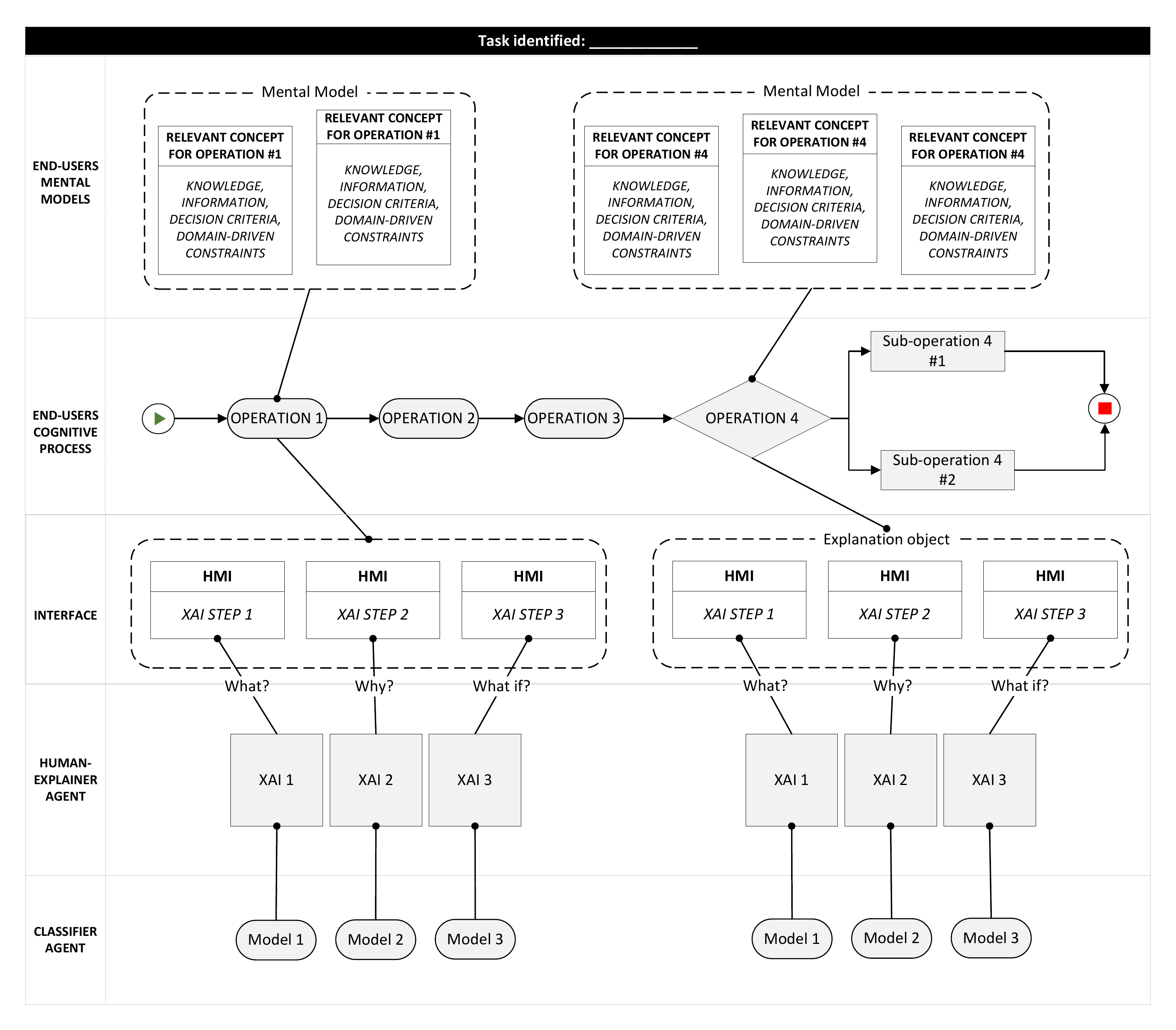}
	\caption{Abstracted Explanation Space}
	\label{fig:space}
\end{figure}

The Abstracted Explanation Space (AES) is a formative tool that aggregates each XAI architecture's components (Figure \ref{fig:space}). The primary function of the AES is to elicit explanation requirements by aligning user-based studies with XAI technical possibilities and capabilities. The tool thereby endorses a boundary object function to achieve common ground with designers, enabling cross-disciplinary design thinking that supplements a narrow algorithmic-centric vision with a socio-technical vision \citep{mueller2021}. Choosing the appropriate format and properties of explanations is facilitated by a clear definition of the end-users needs, tasks (plus the associated mental models), and, more broadly, the work context. It enables designers to turn evidence-based knowledge from user studies into multilayered explanations considering the following factors:
\begin{itemize}
    \item Content: given the cognitive work requirements for a specific task, \textit{what} explanations to present? The sub-goal here is to provide the user with the information that matters only to pursue their work goals. Also, as the \textit{explanation objects} are multilayered, users may query access to delve deeper into the XAI rationale (XAI2 and XAI3) and assess \textit{why} these explanations are trustworthy. 
    \item Format (presentation): explanation display, \textit{how} to present these explanations? Should the explanation be linked to additional information so that users can efficiently process and make sense of the data (e.g., counterfactual explanations)? 
    \item Timing: information flow and temporal aspects, \textit{when} to present these explanations? Understanding the end user's cognitive process helps define temporal specifications.
\end{itemize}

The AES has two distinguishing characteristics: its reliance on user-centered design and ethnographic fieldwork to define XAI structure upon the overarching goals of end-users. Second, its insistence on intertwining end-users characteristics and artificial agents' features defines the explanation space through the user interface (Figure \ref{fig:space}). The end-users are represented by their cognitive process and the mental models associated. The artificial agents are represented by the Human-Agent Explainer and the Classifier Agent. The interface is the channel between the two parties.\\

After defining the task subject to automation, the AES's starting point consists of decomposing the end-users cognitive process and identify the operations that require an explanation (Line 2). In Figure \ref{fig:space}, \textit{Operation 1} and \textit{Operation 4} are the crucial ones for the given task. For each of these operations, the content to be explained is defined by the users' mental models (Line 1). Note that in our hypothetical example, Operation 4 is a knowledge-intensive decision-making task and requires a finer-grained level of decomposition (\textit{sub-operations}). Completing Lines 1 and 2 enable identifying and documenting the explanations to display on the interface for each task/goal pursued (Line 3). The User Interface displays the relevant characteristics of the task from the end-user perspective. The upper part of the AES constitutes the factual user's input that informs designers in selecting the most appropriate explainable models. Moving down the AES’s axis, these factual pieces of evidence are aggregated within the ML solutions: Human-Explainer Agent (Line 4) and Classifier Agent (Line 5). Based on the content that needs to be explained, the ML specialists can judiciously select the most appropriate models and format of explanations regarding each layer of the Human-Explainer Agent (see section \ref{humanexplainer} for more details).

%% file: 4_casestudy.tex
The case study happens in a highly innovative project that is strongly motivated by partially automating aerospace industrial inspection\footnote{\url{https://www.mitacs.ca/en/projects/automated-visual-inspection-sentencing-dressing}}. This manual-extensive and bureaucratic job aims to control the serviceability of high-level aircraft engine components, i.e., certifying the component is free of potentially harmful anomalies. A simplification of the process is presented in the following subsection. The technical solution design involves a consortium of several industrial/academics to develop the hardware (cameras, probes), the software (modules, data model, XAI algorithms), and the human-automation interaction. In the end, some of the inspection process tasks will be automated to assist human operators in human-automation teamwork scenarios. 

\subsection{Targeted Task and Socio-Technical Context}
Aerospace industrial inspection occurs in a highly regulated environment where companies operate within safety policies and international norms. Inspectors are the operators responsible for carefully examining the condition of engine components to determine their serviceability. They do this by examining each component individually for anomalies that could compromise the strength and durability of the part (this phase is also called" visual search"). Then, they interpret the anomaly detected with the work domain's decision-making criteria to decide whether a defect is acceptable, acceptable with a repair, or unacceptable (this phase is also called \textit{decision-making}). These context-dependent criteria come from different sources: operational procedures, domain rules, or result from the knowledge and experience gained by the inspectors. They may relate to the physical characteristics of the defect, the history of the inspected part, its general condition, the possibility of applying subsequent repair operations to remove the defects, or are economical (e.g., avoid unnecessary manufacturing costs and turnaround time).

\subsection{Inspectors Cognitive Process}
To capture the situated aspects of the inspector's decision-making processes, we conducted an ethnographic work analysis. We used various data collection methods involving \textit{in situ} observations (with or without concurrent verbalization), interviews, and field experiments to elicit inspectors' mental models. Understanding how inspectors gather, process, and apply knowledge/information to diagnose component conditions and how they deal with domain-related factors (norms, institutional rules, standard operating procedures). Table \ref{table:method} details the data collection metrics for the user study.\\

\begin{table}[h]
\centering
\footnotesize
\caption{User-inspired study details}
\label{table:method}
\begin{tabular}{m{5cm} m{2cm} m{2cm} c}
\hline
\textbf{Data collection methods} & \textbf{N (inspectors)} & \textbf{Unit of Analysis} & \textbf{Minutes}\\ [0.5 ex]
\hline
Observations (with or without concurrent verbalization) & 7 & 11 & 1110 \\
Interviews & 11 & 26 & 810 \\
Field experiments & 4 & 6 & 660 \\
Total & 11 & 43 & 2580\\
\hline
\end{tabular}
\end{table}

\begin{figure}[h]
	\centering
	\includegraphics[width=1\textwidth]{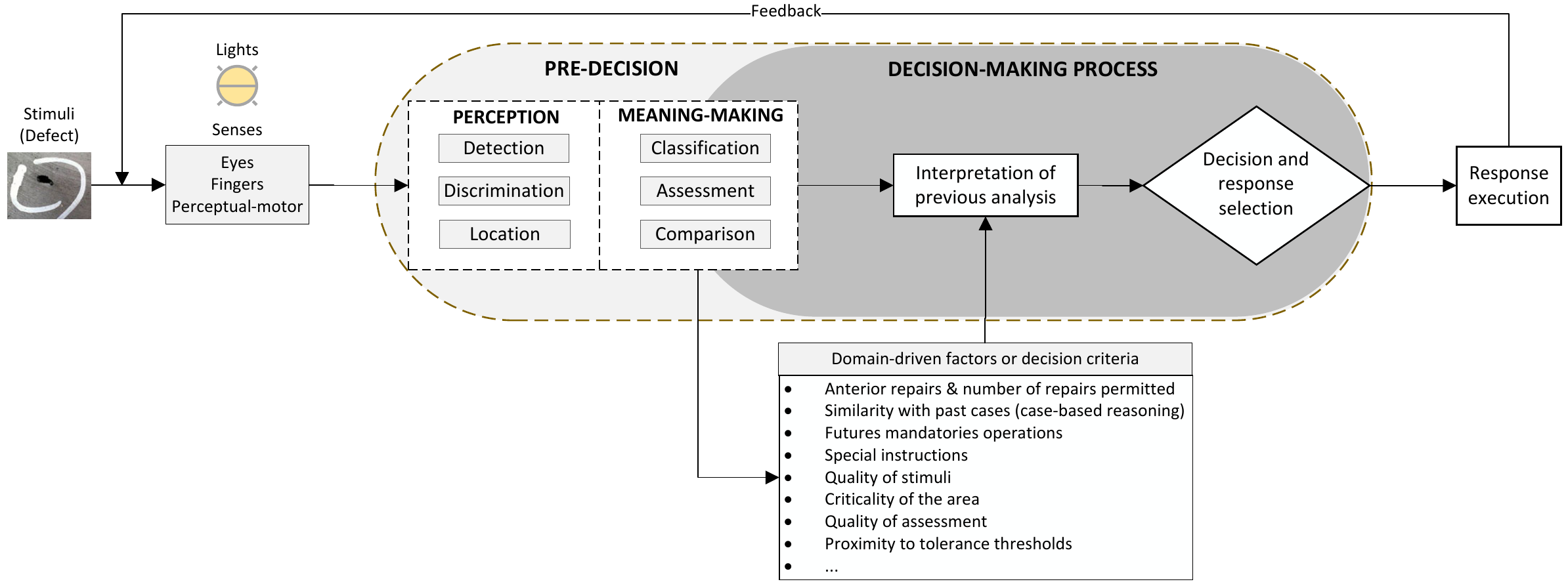}
	\caption{Overall cognitive process of current industrial inspection}
	\label{fig:cognition}
\end{figure}

Inspecting a part is a complex information-processing, meaning-making, and decision-making work (Figure \ref{fig:cognition}). Inspectors' cognitive process (or task-flow) can be unfolded as follows\footnote{In this simplified process, we do not list the upstream tasks that involve the preparation of the workstation. Nor the downstream operations after the execution of the response. See \citep{cabour2021} for a complete description}:
\begin{outline}[enumerate]
	\1 Perception (multi-sensorial search): scan the part to detect any anomaly
	    \2 Detect or discriminate a defect on the part
	    \2 Locate the area where the defect appears
    \1 Meaning-making and decision-making processes: Interpret the anomaly according to the work domain's decision criteria to decide its status
	    \2 \textbf{Classify (characterize) defect type} 
	    \2 Assess the defect characteristics: dimensions (length, width, depth) and the overall condition
	    \2 Compare and interpret defect characteristics with associated rules, standards, norms and domain-driven factors 
	    \2 Decide about defect state: acceptable as is, acceptable with repair or unacceptable
	    \2 Return to the scanning phase until no more anomalies are detected.
    \1 Response execution: actions to be taken following the diagnosis of the defects present on the part
	    \2 Decide about part's state: serviceable, salvageable or non serviceable
	    \2 Prescription of repair operations to remove the defect and restore the part
	    \2 Record defect metrics on a damage record sheet
	    \2 Computer work and optimization
\end{outline}

In our XAI application, we decided to apply the Abstracted Explanation Space (AES) for the first task of the meaning-making and decision-making processes: classification of defect type. Indeed, it is expected that the automated system will carry on this task in future work situations.

\subsection{Uncovering Inspectors Mental Models for the Defect Classification Task}
Analysis of inspectors' work suggests that the relevant knowledge in their mental models for the classification task concerns the distinctive attributes that characterize each defect and the area of defect occurrence.\\

\begin{figure}
	\centering
	\includegraphics[width=0.7\textwidth]{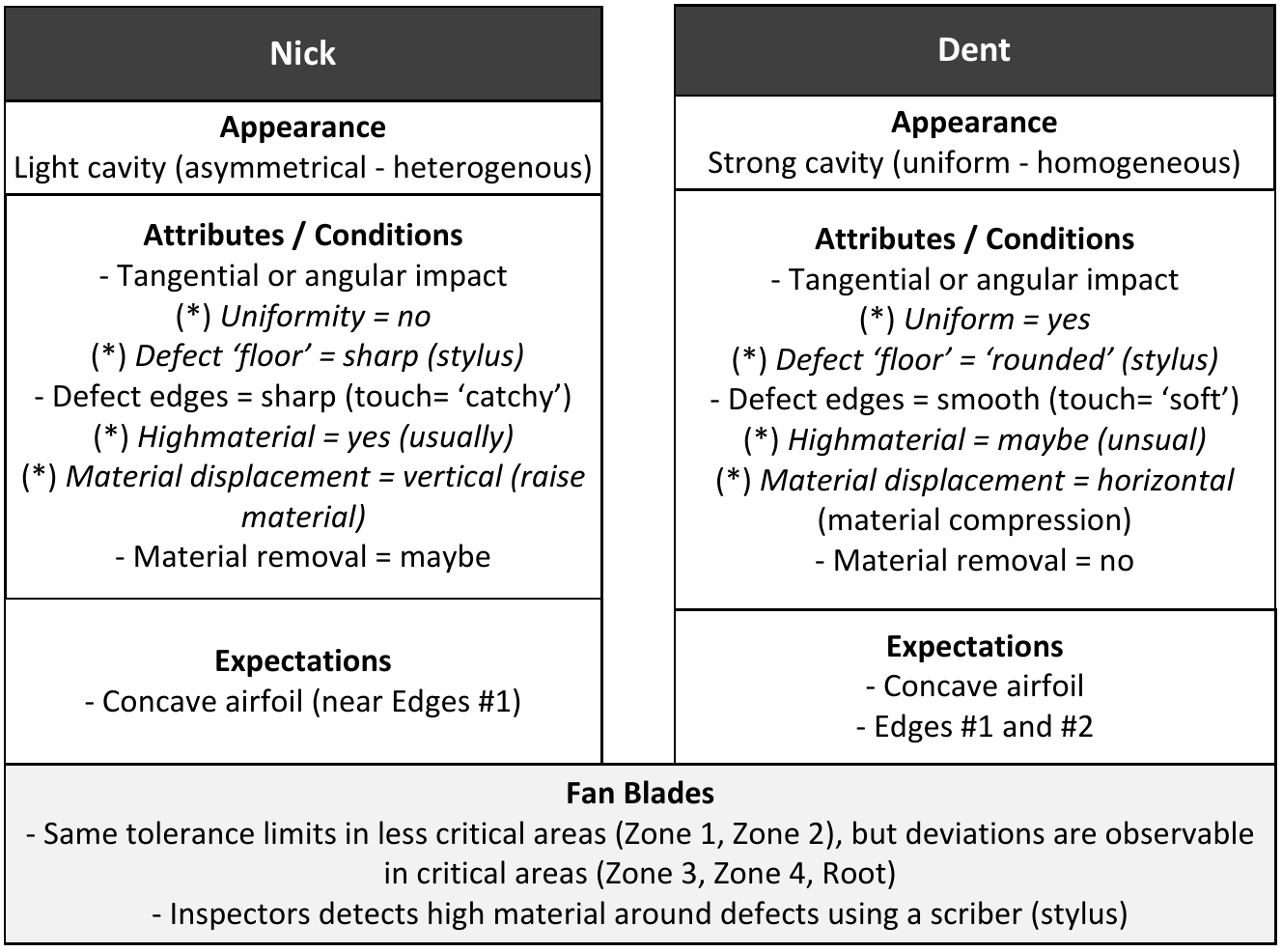}
	\caption{Defect attributes, appearance and expected areas of occurrence for nicks and dents. The attributes in italics with the prefix "(*)" are those that inspectors rely on to distinguish these two similar defects.}
	\label{fig:nickdent}
\end{figure}

\noindent\textbf{Attributes-based classification.} We asked inspectors in semi-structured interviews to define the distinguishing attributes of each class of defect. We also consulted internal documents that defined all defects with associated images and particular attributes. As stated by \cite{cabour2021}, this step is crucial in the inspection process:
\blockquote{\textit{Identifying the type of damage is mandatory, as tolerance thresholds differ according to the type of defect. For the same depth and width, one defect could be acceptable and another one non-acceptable. Inspectors have developed strategies over time to recognize defects by their attributes […] For example, a "nick" is an impact defect that causes a vertical movement of the material along the defect contour (raise material). The presence of high material is distinctive of nicks. Inspectors rely on tactile or perceptual-motor feeling (using a stylus) to detect any material variation on the contour or the bottom (floor) of the defect} (Figure \ref{fig:nickdent}). \textit {A "dent" has similar properties but with a horizontal material displacement. Finding high material around the damage is unusual.}} 

The Figure \ref{fig:nickdent} details the insights gathered through documentation analysis and elicitation of knowledge from the head of inspectors to classify damage types. It provides the relevant cues that guide XAI system specifications. The system should be able to i) \textit{define the relevant combination of attributes that categorize each defect} \citep{cabour2021} ii) provide explanations that allow the inspectors to understand the relevant attributes that led the machine to classify the defect as it is (attribute-based explanations), and iii) generates counterfactual visual explanations for a defect – specifying why the defect was classified as class A rather than class B (counterfactual explanations). During this third step, the Agent explains the decision by specifying the relevant attributes extracted from the inputted images and compare those attributes with similar defects from the same family (nick, dent, gouge, bent, erosion).\\

\begin{figure}[h!]
	\centering
	\includegraphics[width=0.2\textwidth]{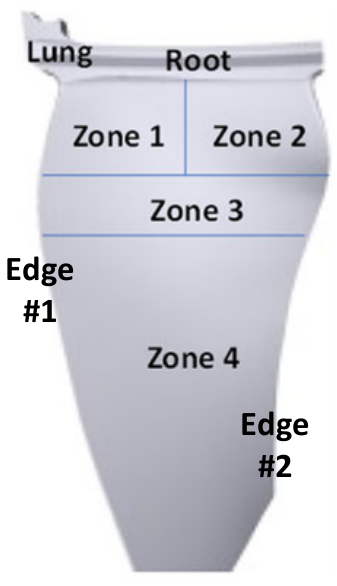}
	\caption{Different part's area for the Fan Blade}
	\label{fig:area}
\end{figure}

\noindent\textbf{Area of defect occurrence (part's location).} Another important point concerns the different areas of the part: Acceptance/refusal criteria change from one to another. For example, the same defect may be acceptable in Zone 1 and not acceptable if it appears in Zone 3 (Figure \ref{fig:area}). The inspector identifies each area with or without additional inspection aid tools and their boundaries. Similarly, the system should have an adequate mapping of the different zones and the boundary between them. Also, inspectors rely on causal reasoning to generate hypotheses about the causes of defects and better identify their class. For example, they expect to see only certain defects in specific areas of the component, depending on their positioning in the engine. The system should indicate the area of defect occurrence to support the causal reasoning of inspectors and allow them to check/trust the accuracy of the system classifications.

\subsection{User interface}
The User Interface materializes the interaction between the inspectors and the artificial agents. Relevant local and global explanations are provided to inspectors for the classification in a usable form. The Human-Explainer Agent provides sample and highlighted images to the end-user. Textual explanations are also generated, as shown in Figure \ref{fig:framework}. Inspectors can request specific explanations from the interface menu. 

\subsection{Human-Explainer Agent}

The Human-Explainer Agent explains the relevant decision made by the Classifier Agent from the end-user perspective. The Classifier Agent performs the detection, location, and classification of the defects, and the Human-Explainer Agent provides meaningful explanations to each of these tasks in parallel.\\

\noindent\textbf{XAI1 Perception - Defect Detection and Location (What? Where?).} Attribute-based explanations for the detection and location of the defect are the first XAI1 component of this process. The Classifier Agent automatically detects and pinpoints the area where a defect is perceived. It is expected that this algorithm had learned that some areas are more sensitive than others. For instance, the detector may know that if a defect is detected in Zone 2, its relevance could be different if found in Zone 4 since Zone 2 needs to meet higher quality standards than Zone 4. However, to ensure that this knowledge is integrated into the process, zone-limits-based decision rules are implemented in the last part of the detector decision. In doing so, the system explains where and in which zone the defect was detected. If a mechanism such as this decision-rules-based mechanism is not implemented, we would have to trust that the detector had learned and intrinsically incorporated this knowledge into its decision mechanism. Thus, we would not have a simple explanation. On the contrary, the Human-Explainer Agent displays a first visual and textual explanation of the defect. So, the inspector can instantly have an explanation of the form "[type of defect] on the [zone]" and the inspected part's image with the defect highlighted.\\

\begin{figure}
\centering
\includegraphics[width=1\textwidth]{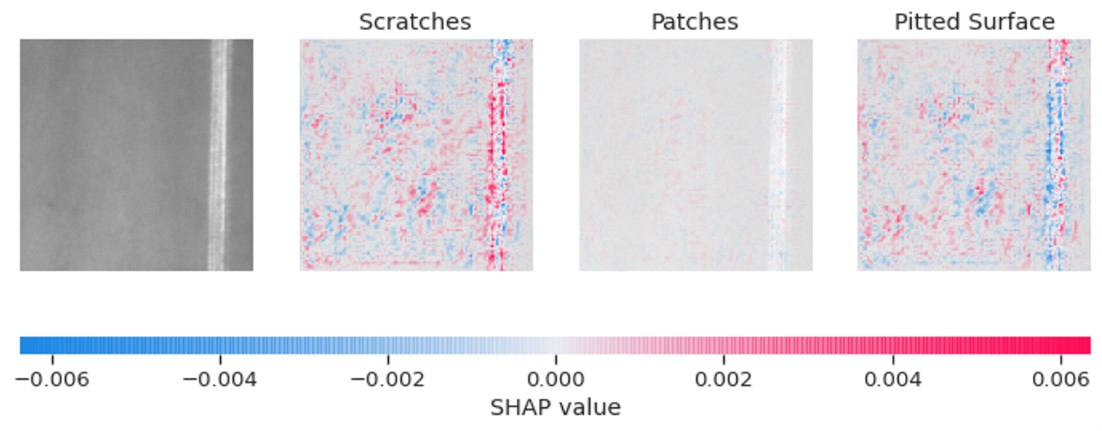}
\caption{Contrastive explanation of a type of defect based on SHAP values. The sample image on the left side is an original image taken from the NEU surface defect database \citep{song2013noise,he2019end,dong2019pga}. The three images left pint point to the pixels that positively affect (red pixels) and negatively (blue points) the classifier's decision.}
\label{fig:shap}
\end{figure}

\noindent\textbf{XAI2 Comprehension - Defect Classification (Why?).} On the other hand, although adversary training and data augmentation are used to train the Classifier Agent, the XAI2 component independently aims to guarantee that if the XAI1 fails, it will do it safely. In this component, adversarial detector methods identify new adversarial observations that XAI1 does not identify.  Methods based on SHAP values or HAWKEYE detectors are used for this purpose. If the inspector validates the observation as an actual adversarial observation, this new observation will feed the example repository of the adversarial detector; thus, it would continuously improve its performance.\\
Also, counterfactual and contrastive local explanations are automatically provided in both visual and textual ways. A set of examples regarding the defect classified is displayed to the inspector to reinforce or change the automatic decision. Figure \ref{fig:shap} shows the explanation on a scratched piece, the red and blue dots on the images, reflecting the pixels' positive and negative relationships with the prediction. In other words, the blue points on the pitted-surface-labeled image, right side of the figure, explains why the classifier does not classify this example as a scratch rather than a pitted surface.  On the contrary, the red points on the same image show the pixels more associated with a pitted surface. Based on this information, the inspector decides whether the defect is a scratch or a pitted surface.\\

\noindent\textbf{XAI3 Projection - Defect Comparison (What if?).} Exposition, sharpening, contrasting, and other pictures' tone parameters are also shifting to reveal some defect characteristics that the XAI2 component may not have detected; this procedure generates hundreds of pictures to be sent back to the classifier and compared with some similar examples. Some textural explanations in the form "if the [photo parameter] is adjusted, the defect would be [type of defect]." This kind of explanation would help, for instance, to compare a nick with a dent; some textures or tone settings may highlight the contrast in the edges of the defect and allow the classifier to detect material displacement. To avoid sending irrelevant information to inspectors, only counter-examples are revealed from the system. On the other hand, since this XAI3 handles parameters initially tuned in the cameras, it might help detect the need to recalibrate those devices or out-of-control environmental factors such as vibrations, noises, and brightness.

\subsection{Classifier Agent}
The Classifier Agent performs the detection, location, and classification of the defects following the inspector's cognitive process. This Agent outputs decisions based on trained ML models and input images. As mentioned above, this Agent also executed adversarial detection to aim for the safety of the system.

\subsection{Abstracted Explanation Space}

\begin{figure}[h]
\centering
\includegraphics[width=1\textwidth]{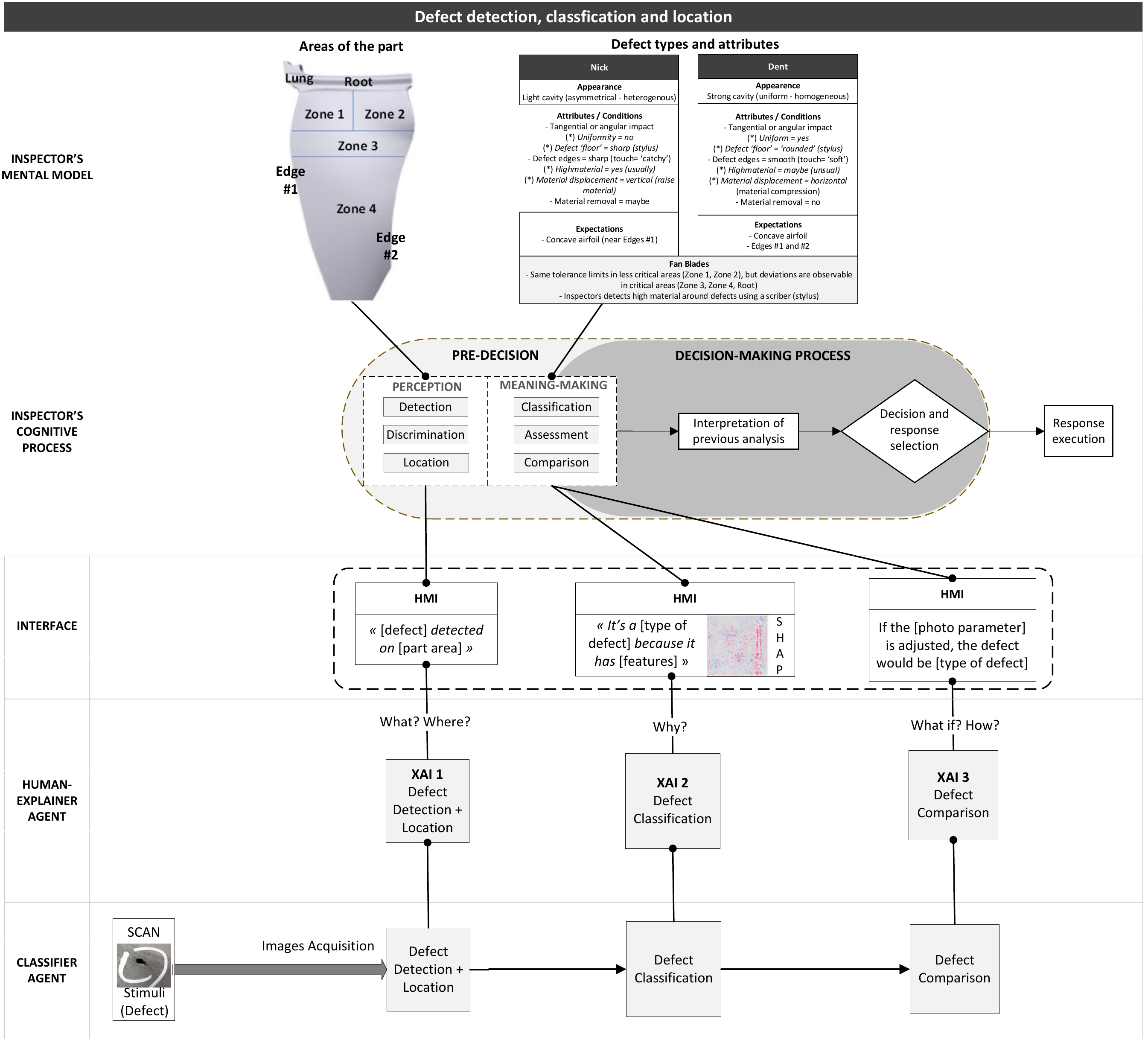}
\caption{Abstracted Explanation Space for defect detection, classification and location tasks in industrial aerospace inspection}
\label{fig:framework}
\end{figure}

Figure \ref{fig:framework} shows the Abstracted Explanation Space that aggregates the different features of our framework for the classification task. Since the automated system will perform the defect detection task using advanced computer vision technology, the sequence of actions will change from the current" manual" process. The inspector will no longer scan the part for potentially dangerous anomalies. The system will be responsible for performing these actions and displaying the information/explanations that the inspector would naturally find during the" manual" process. These elements appear in the first two lines of the AES to detect detection and are implemented using the sequence of XAI1 and XAI2 layers. XAI3 corroborates the output provided by XAI2 in case of unexpected events that surprise the inspector. The system's users may request an explanation if needed.

%% file: 5_dicussion.tex
This paper aimed to introduce a tool that links work context (encompassing the user's needs, tasks, and goals) to the technical development of explanations. We propose a formative tool that integrates past conceptual developments and design principles on user-centered explanations  \citep{mueller2021, ribera2019, sanneman2020, chen2014a}. The ongoing illustrative case study shows the importance of addressing work context and users' needs when designing XAI-based solutions.  For a targeted task, it represents how the relevant components for successful human-XAI interaction are linked together and how the automated system will fit in the envisioned environment. We agree with \cite{mueller2021} on the irrelevance to develop explanations \textit{outside of the work context}. Given that our framework is grounded in user-centered and Human-AI teaming perspectives, the generated explanations should support 1) end-users in accomplishing their operational tasks; and 2) develop shared mental models in the Human-AI team \citep{demir2020}. The pitfall to avoid is providing robust and intelligible explanations that do not align with the pursued objectives. In other words, designing cutting-edge self-explainable technologies that have limited usefulness or perhaps providing an unnecessary amount of information for a specific task. This situation could overload the cognitive state of the inspector, adding information retrieval, processing, and filtering operations on top of the socio-cognitive work to be done. As inspecting a part is a knowledge-intensive work where operators make sense of many data sources, the XAI should effortlessly be integrated into their actual work practices.\\

The AES makes practical contributions from this perspective, as the cognitive task-flow and required knowledge are explicitly represented. Moving down the AES axes, ML specialists can define the level of explanation needed (partial/complete, local/global) and the desirable approaches to achieve it (e.g., human-explainer mechanism). To the best of our knowledge, this is the first work that aims to develop a practical tool/framework that bridges the two areas necessary for explanation design, user studies and XAI system design, and with a demonstrative case study in the aerospace domain. The tool can also be extended to encompass additional actors in the socio-technical context who will interact directly or indirectly with the technical device. As the introduction of technological tools changes work dynamics, other socio-technical actors (more or less distant from the device's physical location) may need explanations or understanding of XAI system rationale. Staying within the context of our case study, diagnosing the condition of a part as serviceable or unserviceable has implications for the work of inspectors and technical representatives. They must inform customers that a part needs to be changed and find a compromise with them. To do this, they may need additional information (explanations) that the inspectors would provide in the manual process, e.g., reason(s) for rejecting the part, instructions for its replacement. To explore this aspect, we plan to apply the framework to other relevant inspector tasks, especially those crucial to the decision-making process. In addition, it might be interesting to use the AES for other socio-technical actors.\\

We recognize the limitations of this work in demonstrating how the effectiveness of the AES can be measured, particularly the user-perceived quality factors \citep{nunes2017}. This paper focuses on defining the explanation space, and we plan to validate its utility by conducting evaluation-based researches with relevant quantitative/qualitative measures. The metrics will encompass the performance of the technical system, the human, and the human-AI teamwork during simulation-informed activities. Also, the formative tool's usefulness must be defined according to how it is applied in an innovation process, for instance, how the artifact supports cross-functional work. Additional case studies may be needed to produce this type of empirical evaluation.

We also recognize that the current proposal may evolve in terms of additional layers or modification of existing ones. For example, new forms of advanced human-machine interactions are emerging with haptic devices and natural language communication. These new technologies offer new possibilities that could change the third line of the AES: \textit{interface}. This line represents the communication channel between users and the XAI, regardless of the interaction modality chosen. Thereby, we contend that the five pillars of the framework constitute the foundation of the explanation space.\\

To conclude, this paper introduced a user-centered framework for eXplainable AI (XAI) with its associated features and illustrated it using the case of aerospace industrial inspection. The framework relies on the Abstracted Explanation Space (AES): a formative tool that defines an explanation space for designers of future XAI-based systems to aggregate social (end-users) and technical views when designing explanations. Indeed, little attention has been paid to the users who will receive the explanations and how they will collaborate with artificial agents in operational environments. This paper thereby offers the following contributions: 1) a proposition of a framework to systematically articulates the explanations that end-users require for a given task with the XAI features to develop; 2) a presentation of the AES that allows cross-disciplinary dialogue between designers to surface the types of models and level of explanation to align with users' goals and tasks; 3) a demonstration of the framework/tool in a real-world application in the aerospace domain. In this way, the paper advances practical knowledge and interdisciplinary endeavors to shape technology development to meet human and societal needs.\\

%% file: 0_main.bbl
\begin{thebibliography}{10}

\bibitem{vasey2021}
B.~Vasey, D.~A. Clifton, G.~S. Collins, A.~K. Denniston, L.~Faes, B.~F. Geerts,
  X.~Liu, L.~Morgan, P.~Watkinson, P.~McCulloch, and {The DECIDE-AI Steering
  Group}, ``{{DECIDE}}-{{AI}}: New reporting guidelines to bridge the
  development-to-implementation gap in clinical artificial intelligence,'' {\em
  Nature Medicine}, vol.~27, pp.~186--187, Feb. 2021.

\bibitem{cabitza2019}
F.~Cabitza and J.-D. Zeitoun, ``The proof of the pudding: In praise of a
  culture of real-world validation for medical artificial intelligence,'' {\em
  Annals of Translational Medicine}, vol.~7, no.~8, p.~9, 2019.

\bibitem{jiao_towards_2020}
J.~R. Jiao, F.~Zhou, N.~Z. Gebraeel, and V.~Duffy, ``Towards augmenting
  cyber-physical-human collaborative cognition for human-automation interaction
  in complex manufacturing and operational environments,'' {\em International
  Journal of Production Research}, vol.~0, pp.~1--23, Mar. 2020.
\newblock Publisher: Taylor \& Francis \_eprint:
  https://doi.org/10.1080/00207543.2020.1722324.

\bibitem{peeters2021}
M.~M. Peeters, J.~van Diggelen, K.~Van Den~Bosch, A.~Bronkhorst, M.~A.
  Neerincx, J.~M. Schraagen, and S.~Raaijmakers, ``Hybrid collective
  intelligence in a human--ai society,'' {\em AI \& SOCIETY}, vol.~36, no.~1,
  pp.~217--238, 2021.

\bibitem{cabour2021}
G.~Cabour, {\'E}.~Ledoux, and S.~Bassetto, ``A {{Work}}-{{Centered Approach}}
  for {{Cyber}}-{{Physical}}-{{Social System Design}}: {{Applications}} in
  {{Aerospace Industrial Inspection}},'' {\em arXiv:2101.05385 [cs]}, Jan.
  2021.

\bibitem{cutillo2020}
C.~M. Cutillo, K.~R. Sharma, L.~Foschini, S.~Kundu, M.~Mackintosh, and K.~D.
  Mandl, ``Machine intelligence in healthcare\textemdash perspectives on
  trustworthiness, explainability, usability, and transparency,'' {\em npj
  Digital Medicine}, vol.~3, pp.~1--5, Mar. 2020.

\bibitem{johnson2017}
M.~Johnson, J.~M. Bradshaw, and P.~J. Feltovich, ``Tomorrow's
  {{Human}}\textendash{{Machine Design Tools}}: {{From Levels}} of
  {{Automation}} to {{Interdependencies}}:,'' {\em Journal of Cognitive
  Engineering and Decision Making}, Oct. 2017.

\bibitem{doshi-velez2017}
F.~{Doshi-Velez} and B.~Kim, ``Towards {{A Rigorous Science}} of
  {{Interpretable Machine Learning}},'' {\em arXiv:1702.08608 [cs, stat]}, Mar.
  2017.

\bibitem{rajabiyazdi2020}
F.~Rajabiyazdi and G.~A. Jamieson, ``A review of transparency (seeing-into)
  models,'' in {\em 2020 IEEE International Conference on Systems, Man, and
  Cybernetics (SMC)}, pp.~302--308, IEEE, 2020.

\bibitem{mueller2019}
S.~T. Mueller, R.~R. Hoffman, W.~Clancey, A.~Emrey, and G.~Klein, ``Explanation
  in {{Human}}-{{AI Systems}}: {{A Literature Meta}}-{{Review}}, {{Synopsis}}
  of {{Key Ideas}} and {{Publications}}, and {{Bibliography}} for {{Explainable
  AI}},'' {\em arXiv:1902.01876 [cs]}, Feb. 2019.

\bibitem{mueller2021}
S.~T. Mueller, E.~S. Veinott, R.~R. Hoffman, G.~Klein, L.~Alam, T.~Mamun, and
  W.~J. Clancey, ``Principles of {{Explanation}} in {{Human}}-{{AI Systems}},''
  {\em arXiv:2102.04972 [cs]}, Feb. 2021.

\bibitem{hoffman2018}
R.~R. Hoffman, S.~T. Mueller, G.~Klein, and J.~Litman, ``Metrics for
  explainable {{AI}}: {{Challenges}} and prospects,'' {\em arXiv preprint
  arXiv:1812.04608}, 2018.

\bibitem{johnson2021}
M.~Johnson and J.~M. Bradshaw, ``The role of interdependence in trust,'' in
  {\em Trust in Human-Robot Interaction}, pp.~379--403, Elsevier, 2021.

\bibitem{miller2019}
T.~Miller, ``Explanation in artificial intelligence: Insights from the social
  sciences,'' {\em Artificial intelligence}, vol.~267, pp.~1--38, 2019.

\bibitem{ribera2019}
M.~Ribera and A.~Lapedriza, ``Can we do better explanations? {{A}} proposal of
  {{User}}-{{Centered Explainable AI}},'' {\em Los Angeles}, p.~7, 2019.

\bibitem{selbst2019}
A.~D. Selbst, D.~Boyd, S.~A. Friedler, S.~Venkatasubramanian, and J.~Vertesi,
  ``Fairness and {{Abstraction}} in {{Sociotechnical Systems}},'' in {\em
  Proceedings of the {{Conference}} on {{Fairness}}, {{Accountability}}, and
  {{Transparency}}}, {{FAT}}* '19, ({New York, NY, USA}), pp.~59--68,
  {Association for Computing Machinery}, Jan. 2019.

\bibitem{chen2014a}
J.~Y. Chen and M.~J. Barnes, ``Human--agent teaming for multirobot control: A
  review of human factors issues,'' {\em IEEE Transactions on Human-Machine
  Systems}, vol.~44, no.~1, pp.~13--29, 2014.

\bibitem{seeber2020m}
I.~Seeber, E.~Bittner, R.~O. Briggs, T.~De~Vreede, G.-J. De~Vreede, A.~Elkins,
  R.~Maier, A.~B. Merz, S.~Oeste-Rei{\ss}, N.~Randrup, {\em et~al.}, ``Machines
  as teammates: A research agenda on ai in team collaboration,'' {\em
  Information \& management}, vol.~57, no.~2, p.~103174, 2020.

\bibitem{matthews2021}
G.~Matthews, A.~R. Panganiban, J.~Lin, M.~Long, and M.~Schwing,
  ``Super-machines or sub-humans: Mental models and trust in intelligent
  autonomous systems,'' in {\em Trust in Human-Robot Interaction}, pp.~59--82,
  Elsevier, 2021.

\bibitem{clancey1983}
W.~J. Clancey, ``The epistemology of a rule-based expert system \textemdash a
  framework for explanation,'' {\em Artificial Intelligence}, vol.~20,
  pp.~215--251, May 1983.

\bibitem{moore1988}
J.~Moore, ``Explanation in {{Expert Systems}} : {{A Survey}},'' 1988.

\bibitem{zouinar2020}
M.~Zouinar, ``{\'Evolutions de l'Intelligence Artificielle : quels enjeux pour
  l'activit\'e humaine et la relation Humain-Machine au travail ?},'' {\em
  Activit\'es}, Apr. 2020.

\bibitem{roth1987}
E.~M. Roth, K.~B. Bennett, and D.~D. Woods, ``Human interaction with an
  ``intelligent'' machine,'' {\em International Journal of Man-Machine
  Studies}, vol.~27, pp.~479--525, Nov. 1987.

\bibitem{lewis2021deep}
M.~Lewis, H.~Li, and K.~Sycara, ``Deep learning, transparency, and trust in
  human robot teamwork,'' in {\em Trust in Human-Robot Interaction},
  pp.~321--352, Elsevier, 2021.

\bibitem{habli2020}
I.~Habli, T.~Lawton, and Z.~Porter, ``Artificial intelligence in health care:
  accountability and safety,'' {\em Bulletin of the World Health Organization},
  vol.~98, no.~4, p.~251, 2020.

\bibitem{shmelova2020}
T.~Shmelova, A.~Sterenharz, and S.~Dolgikh, ``Artificial intelligence in
  aviation industries: Methodologies, education, applications, and
  opportunities,'' in {\em Handbook of Research on Artificial Intelligence
  Applications in the Aviation and Aerospace Industries}, pp.~1--35, IGI
  Global, 2020.

\bibitem{mnih2015human}
V.~Mnih, K.~Kavukcuoglu, D.~Silver, A.~A. Rusu, J.~Veness, M.~G. Bellemare,
  A.~Graves, M.~Riedmiller, A.~K. Fidjeland, G.~Ostrovski, {\em et~al.},
  ``Human-level control through deep reinforcement learning,'' {\em nature},
  vol.~518, no.~7540, pp.~529--533, 2015.

\bibitem{marcus2019rebooting}
G.~Marcus and E.~Davis, {\em Rebooting AI: Building artificial intelligence we
  can trust}.
\newblock Pantheon, 2019.

\bibitem{lapuschkin2019}
S.~Lapuschkin, S.~W{\"a}ldchen, A.~Binder, G.~Montavon, W.~Samek, and K.-R.
  M{\"u}ller, ``Unmasking {{Clever Hans}} predictors and assessing what
  machines really learn,'' {\em Nature Communications}, vol.~10, p.~1096, Mar.
  2019.

\bibitem{akatsuka_illuminating_2019}
J.~Akatsuka, Y.~Yamamoto, T.~Sekine, Y.~Numata, H.~Morikawa, K.~Tsutsumi,
  M.~Yanagi, Y.~Endo, H.~Takeda, T.~Hayashi, M.~Ueki, G.~Tamiya, I.~Maeda,
  M.~Fukumoto, A.~Shimizu, T.~Tsuzuki, G.~Kimura, and Y.~Kondo, ``Illuminating
  {Clues} of {Cancer} {Buried} in {Prostate} {MR} {Image}: {Deep} {Learning}
  and {Expert} {Approaches},'' {\em Biomolecules}, vol.~9, Oct. 2019.

\bibitem{tomsett_why_2018}
R.~Tomsett, A.~Widdicombe, T.~Xing, S.~Chakraborty, S.~Julier, P.~Gurram,
  R.~Rao, and M.~Srivastava, ``Why the {Failure}? {How} {Adversarial}
  {Examples} {Can} {Provide} {Insights} for {Interpretable} {Machine}
  {Learning},'' in {\em 2018 21st {International} {Conference} on {Information}
  {Fusion} ({FUSION})}, pp.~838--845, July 2018.

\bibitem{lakkaraju2016interpretable}
H.~Lakkaraju, S.~H. Bach, and J.~Leskovec, ``Interpretable decision sets: A
  joint framework for description and prediction,'' in {\em Proceedings of the
  22nd ACM SIGKDD international conference on knowledge discovery and data
  mining}, pp.~1675--1684, 2016.

\bibitem{pekka2018european}
A.~Pekka, W.~Bauer, U.~Bergmann, M.~Bielikov{\'a}, C.~Bonefeld-Dahl, Y.~Bonnet,
  L.~Bouarfa, {\em et~al.}, ``The european commission’s high-level expert
  group on artificial intelligence: Ethics guidelines for trustworthy ai,''
  {\em Working Document for stakeholders’ consultation. Brussels}, pp.~1--37,
  2018.

\bibitem{yeung2020recommendation}
K.~Yeung, ``Recommendation of the council on artificial intelligence (oecd),''
  {\em International Legal Materials}, vol.~59, no.~1, pp.~27--34, 2020.

\bibitem{xiao_generating_2019}
C.~Xiao, B.~Li, J.-Y. Zhu, W.~He, M.~Liu, and D.~Song, ``Generating
  {Adversarial} {Examples} with {Adversarial} {Networks},'' {\em
  arXiv:1801.02610 [cs, stat]}, Feb. 2019.
\newblock arXiv: 1801.02610.

\bibitem{morales2019case}
A.~Morales-Forero and S.~Bassetto, ``Case study: A semi-supervised methodology
  for anomaly detection and diagnosis,'' in {\em 2019 IEEE International
  Conference on Industrial Engineering and Engineering Management (IEEM)},
  pp.~1031--1037, IEEE, 2019.

\bibitem{fidel_when_2019}
G.~Fidel, R.~Bitton, and A.~Shabtai, ``When {Explainability} {Meets}
  {Adversarial} {Learning}: {Detecting} {Adversarial} {Examples} using {SHAP}
  {Signatures},'' {\em arXiv:1909.03418 [cs, stat]}, Sept. 2019.
\newblock arXiv: 1909.03418.

\bibitem{tramer_adversarial_2019}
F.~Tramer and D.~Boneh, ``Adversarial training and robustness for multiple
  perturbations,'' {\em arXiv preprint arXiv:1904.13000}, 2019.

\bibitem{mor-yosef_ranking_1990}
S.~Mor-Yosef, A.~Samueloff, B.~Modan, D.~Navot, and J.~G. Schenker, ``Ranking
  the risk factors for cesarean: logistic regression analysis of a nationwide
  study,'' {\em Obstetrics and Gynecology}, vol.~75, pp.~944--947, June 1990.

\bibitem{kobrin2011investigation}
J.~L. Kobrin, S.~Sinharay, S.~J. Haberman, and M.~Chajewski, ``An investigation
  of the fit of linear regression models to data from an sat{\textregistered}
  validity study,'' {\em ETS Research Report Series}, vol.~2011, no.~1,
  pp.~i--21, 2011.

\bibitem{rudin2019stop}
C.~Rudin, ``Stop explaining black box machine learning models for high stakes
  decisions and use interpretable models instead,'' {\em Nature Machine
  Intelligence}, vol.~1, no.~5, pp.~206--215, 2019.

\bibitem{lundberg2017unified}
S.~M. Lundberg and S.-I. Lee, ``A unified approach to interpreting model
  predictions,'' in {\em Advances in neural information processing systems},
  pp.~4765--4774, 2017.

\bibitem{ribeiro2016should}
M.~T. Ribeiro, S.~Singh, and C.~Guestrin, ``"why should i trust you?"
  explaining the predictions of any classifier,'' in {\em Proceedings of the
  22nd ACM SIGKDD international conference on knowledge discovery and data
  mining}, pp.~1135--1144, 2016.

\bibitem{konig2008g}
R.~Konig, U.~Johansson, and L.~Niklasson, ``G-rex: A versatile framework for
  evolutionary data mining,'' in {\em 2008 IEEE International Conference on
  Data Mining Workshops}, pp.~971--974, IEEE, 2008.

\bibitem{henelius2017interpreting}
A.~Henelius, K.~Puolam{\"a}ki, and A.~Ukkonen, ``Interpreting classifiers
  through attribute interactions in datasets,'' {\em arXiv preprint
  arXiv:1707.07576}, 2017.

\bibitem{rathi2019generating}
S.~Rathi, ``Generating counterfactual and contrastive explanations using
  shap,'' {\em arXiv preprint arXiv:1906.09293}, 2019.

\bibitem{avserivskis2014gamification}
D.~A{\v{s}}eri{\v{s}}kis and R.~Dama{\v{s}}evi{\v{c}}ius, ``Gamification of a
  project management system,'' in {\em Proc. of Int. Conference on Advances in
  Computer-Human Interactions ACHI2014}, pp.~200--207, Citeseer, 2014.

\bibitem{shahri2014towards}
A.~Shahri, M.~Hosseini, K.~Phalp, J.~Taylor, and R.~Ali, ``Towards a code of
  ethics for gamification at enterprise,'' in {\em IFIP working conference on
  the practice of enterprise modeling}, pp.~235--245, Springer, 2014.

\bibitem{naiseh2020}
M.~Naiseh, N.~Jiang, J.~Ma, and R.~Ali, ``Personalising explainable
  recommendations: literature and conceptualisation,'' in {\em World Conference
  on Information Systems and Technologies}, pp.~518--533, Springer, 2020.

\bibitem{morales2021un}
A.~Morales-Forero, S.~Bassetto, and E.~Coatanea, ``Ai interpretable models for
  high-stake decisions based on concepts and context.'' unpublished, 2021.

\bibitem{mohseni_multidisciplinary_2020}
S.~Mohseni, N.~Zarei, and E.~D. Ragan, ``A {Multidisciplinary} {Survey} and
  {Framework} for {Design} and {Evaluation} of {Explainable} {AI} {Systems},''
  {\em arXiv:1811.11839 [cs]}, Apr. 2020.
\newblock arXiv: 1811.11839.

\bibitem{sanneman2020}
L.~Sanneman and J.~A. Shah, ``A {{Situation Awareness}}-{{Based Framework}} for
  {{Design}} and {{Evaluation}} of {{Explainable AI}},'' in {\em Explainable,
  {{Transparent Autonomous Agents}} and {{Multi}}-{{Agent Systems}}}
  (D.~Calvaresi, A.~Najjar, M.~Winikoff, and K.~Fr{\"a}mling, eds.), ({Cham}),
  pp.~94--110, {Springer International Publishing}, 2020.

\bibitem{crabtree2012}
A.~Crabtree, M.~Rouncefield, and P.~Tolmie, ``Informing design,'' in {\em Doing
  Design Ethnography}, pp.~135--158, Springer, 2012.

\bibitem{bisantz2015}
A.~Bisantz, E.~M. Roth, and J.~{Watts-Englert}, ``Study and {{Analysis}} of
  {{Complex Cognitive Work}},'' in {\em Evaluation of {{Human Work}}} (J.~R.
  Wilson and S.~Sharples, eds.), pp.~61--82, {CRC Press}, May 2015.

\bibitem{zsambok2014}
C.~E. Zsambok and G.~Klein, {\em Naturalistic decision making}.
\newblock Psychology Press, 2014.

\bibitem{klein2016}
G.~Klein and C.~Wright, ``Macrocognition: from theory to toolbox,'' {\em
  Frontiers in psychology}, vol.~7, p.~54, 2016.

\bibitem{vicente1999}
K.~J. Vicente, {\em Cognitive work analysis: Toward safe, productive, and
  healthy computer-based work}.
\newblock CRC Press, 1999.

\bibitem{st-vincent2014}
{St-Vincent}, V{\'e}zina, Bellemare, Denis, Ledoux, and Imbeau, {\em Ergonomic
  {{Intervention}}}.
\newblock {Institut de recherche Robert-Sauv\'e en sant\'e et en s\'ecurit\'e
  du travail}, 2014.

\bibitem{salembier2021}
P.~Salembier and I.~Wagner, ``Studies of work ‘in the wild’,'' {\em
  Computer Supported Cooperative Work (CSCW)}, pp.~1--20, 2021.

\bibitem{endsley2007}
M.~R. Endsley, R.~Hoffman, D.~Kaber, and E.~Roth, ``Cognitive engineering and
  decision making: An overview and future course,'' {\em Journal of Cognitive
  Engineering and Decision Making}, vol.~1, no.~1, pp.~1--21, 2007.

\bibitem{cabour2021a}
G.~Cabour, {\'E}.~Ledoux, and S.~Bassetto, ``Extending {{System Performance
  Past}} the {{Boundaries}} of {{Technical Maturity}}: {{Human}}-{{Agent
  Teamwork Perspective}} for {{Industrial Inspection}},'' {\em arXiv:2102.03241
  [cs]}, Feb. 2021.

\bibitem{goh2020}
Y.~M. Goh, S.~Micheler, A.~Sanchez-Salas, K.~Case, D.~Bumblauskas, and
  R.~Monfared, ``A variability taxonomy to support automation decision-making
  for manufacturing processes,'' {\em Production Planning \& Control}, vol.~31,
  no.~5, pp.~383--399, 2020.

\bibitem{shadbolt2015}
N.~R. Shadbolt, P.~R. Smart, J.~Wilson, and S.~Sharples, ``Knowledge
  elicitation,'' {\em Evaluation of human work}, pp.~163--200, 2015.

\bibitem{milton2007}
N.~R. Milton, {\em Knowledge acquisition in practice: a step-by-step guide}.
\newblock Springer Science \& Business Media, 2007.

\bibitem{wilson2015}
J.~R. Wilson and S.~Sharples, {\em Evaluation of {{Human Work}}}.
\newblock {CRC Press}, Apr. 2015.

\bibitem{nickerson2013}
R.~C. Nickerson, U.~Varshney, and J.~Muntermann, ``A method for taxonomy
  development and its application in information systems,'' {\em European
  Journal of Information Systems}, vol.~22, no.~3, pp.~336--359, 2013.

\bibitem{mcmeekin2020}
N.~McMeekin, O.~Wu, E.~Germeni, and A.~Briggs, ``How methodological frameworks
  are being developed: evidence from a scoping review,'' {\em BMC medical
  research methodology}, vol.~20, no.~1, pp.~1--9, 2020.

\bibitem{friedman2018}
S.~Friedman, K.~Forbus, and B.~Sherin, ``Representing, running, and revising
  mental models: A computational model,'' {\em Cognitive science}, vol.~42,
  no.~4, pp.~1110--1145, 2018.

\bibitem{elsawah2015}
S.~Elsawah, J.~H. Guillaume, T.~Filatova, J.~Rook, and A.~J. Jakeman, ``A
  methodology for eliciting, representing, and analysing stakeholder knowledge
  for decision making on complex socio-ecological systems: from cognitive maps
  to agent-based models,'' {\em Journal of environmental management}, vol.~151,
  pp.~500--516, 2015.

\bibitem{shepherd2015}
A.~Shepherd and R.~B. Stammers, ``Task analysis,'' in {\em Evaluation of
  {{Human Work}}} (J.~R. Wilson and S.~Sharples, eds.), pp.~139--162, {CRC
  Press}, May 2015.

\bibitem{chen2014b}
J.~Y. Chen, K.~Procci, M.~Boyce, J.~Wright, A.~Garcia, and M.~Barnes,
  ``Situation awareness-based agent transparency,'' tech. rep., Army research
  lab aberdeen proving ground md human research and engineering~…, 2014.

\bibitem{preece2018}
A.~D. Preece, D.~Harborne, D.~Braines, R.~Tomsett, and S.~Chakraborty,
  ``Stakeholders in explainable {AI},'' {\em CoRR}, vol.~abs/1810.00184, 2018.

\bibitem{song2013noise}
K.~Song and Y.~Yan, ``A noise robust method based on completed local binary
  patterns for hot-rolled steel strip surface defects,'' {\em Applied Surface
  Science}, vol.~285, pp.~858--864, 2013.

\bibitem{he2019end}
Y.~He, K.~Song, Q.~Meng, and Y.~Yan, ``An end-to-end steel surface defect
  detection approach via fusing multiple hierarchical features,'' {\em IEEE
  Transactions on Instrumentation and Measurement}, vol.~69, no.~4,
  pp.~1493--1504, 2019.

\bibitem{dong2019pga}
H.~Dong, K.~Song, Y.~He, J.~Xu, Y.~Yan, and Q.~Meng, ``Pga-net: Pyramid feature
  fusion and global context attention network for automated surface defect
  detection,'' {\em IEEE Transactions on Industrial Informatics}, vol.~16,
  no.~12, pp.~7448--7458, 2019.

\bibitem{demir2020}
M.~Demir, N.~J. McNeese, and N.~J. Cooke, ``Understanding human-robot teams in
  light of all-human teams: Aspects of team interaction and shared cognition,''
  {\em International Journal of Human-Computer Studies}, vol.~140, p.~102436,
  2020.

\bibitem{nunes2017}
I.~Nunes and D.~Jannach, ``A systematic review and taxonomy of explanations in
  decision support and recommender systems,'' {\em User Modeling and
  User-Adapted Interaction}, vol.~27, no.~3, pp.~393--444, 2017.

\end{thebibliography}
